\documentclass[review]{elsarticle}

\makeatletter
\def\ps@pprintTitle{
	\def\@oddfoot{\centerline{\thepage}}
}
\makeatother

\usepackage{hyperref}
\usepackage{mathtools}
\usepackage{gensymb}
\usepackage{graphicx}
\graphicspath{{images/}}
\usepackage{booktabs}
\usepackage{float}
\usepackage{listings}
\usepackage{multirow}

\journal{Future Generation Computer Systems}

\begin{document}
\begin{titlepage}
\noindent \textbf{IoFClime: The fuzzy logic and the Internet of Things to control indoor temperature regarding the outdoor ambient conditions}\\

\noindent \textbf{Notice:} this is the author's version of a work accepted to be published in Future Generation Computer Systems. It is posted here for your personal use and following the Elsevier copyright policies. Changes resulting from the publishing process, such as editing, corrections, structural formatting, and other quality control mechanisms may not be reflected in this document. A more definitive version can be consulted on:\\

\noindent Meana-Llori\'an, D., Gonz\'alez Garc\'ia, C., Pelayo G-Bustelo, B. C., Cueva Lovelle, J. M., \& Garcia-Fernandez, N. (2016). IoFClime: The fuzzy logic and the Internet of Things to control indoor temperature regarding the outdoor ambient conditions. \textit{Future Generation Computer Systems}. \url{https://doi.org/10.1016/j.future.2016.11.020}\\

\noindent \textbf{\copyright{} 2016.} This manuscript version is made available under the CC-BY-NC-ND 4.0 license \url{http://creativecommons.org/licenses/by-nc-nd/4.0/}
\end{titlepage}

\begin{frontmatter}

\title{IoFClime: The fuzzy logic and the Internet of Things to control indoor temperature regarding the outdoor ambient conditions}

\author[uniovi]{Daniel Meana-Llori\'an\corref{cor1}}
\ead{danielmeanallorian@gmail.com}

\author[uniovi]{Cristian Gonz\'alez Garc\'ia}
\ead{gonzalezgarciacristian@hotmail.com}

\author[uniovi]{B. Cristina Pelayo G-Bustelo}
\ead{crispelayo@uniovi.es}

\author[uniovi]{Juan Manuel Cueva Lovelle}
\ead{cueva@uniovi.es}

\author[uniovi]{Nestor Garcia-Fernandez}
\ead{nestor@uniovi.es}

\cortext[cor1]{Corresponding author}

\address[uniovi]{University of Oviedo, Department of Computer Science, Sciences Building, C/Calvo Sotelo s/n 33007, Oviedo, Asturias, Spain}

\begin{abstract}
The Internet of Things is arriving to our homes or cities through fields already known like Smart Homes, Smart Cities, or Smart Towns. The monitoring of environmental conditions of cities can help to adapt the indoor locations of the cities in order to be more comfortable for people who stay there. A way to improve the indoor conditions is an efficient temperature control, however, it depends on many factors like the different combinations of outdoor temperature and humidity. Therefore, adjusting the indoor temperature is not setting a value according to other value. There are many more factors to take into consideration, hence the traditional logic based in binary states cannot be used. Many problems cannot be solved with a set of binary solutions and we need a new way of development. Fuzzy logic is able to interpret many states, more than two states, giving to computers the capacity to react in a similar way to people. In this paper we will propose a new approach to control the temperature using the Internet of Things together its platforms and fuzzy logic regarding not only the indoor temperature but also the outdoor temperature and humidity in order to save energy and to set a more comfortable environment for their users. Finally, we will conclude that the fuzzy approach allows us to achieve an energy saving around 40\% and thus, save money.
\end{abstract}


\end{frontmatter}

\section{Introduction}
The Internet of Things (IoT) is a term very popular nowadays. Every day we can listen people taking about Smart Homes \cite{gu2009content, hribernik2011co}, Smart Cities \cite{hao2012application, lea2014smart}, Smart Earth \cite{hao2012application}, and many other kinds of distributed intelligence around heterogeneous and ubiquitous objects. The IoT allows gathering a huge quantity of data that can be processed to help making different decisions. These data may be very varied and confused and processing them might become inoperable.

Humans and computers make decisions in a different way. Whereas human reason uses words, computers use numbers \cite{nikravesh2007evolution}. Moreover, even though the logic applied by humans seems more primitive, they can make better decisions in the real-world when an unexpected problem appears. Computing with words could improve the capability of computers to deal with problems of real-world and thus, improving decision making \cite{nikravesh2007evolution}. 

The human capability to take decisions without computations is usually referred as `common sense'. Common sense allows us to take decisions quickly although they are not always the best ones. For example, in the past, we thought that the Earth was flat due to common sense \cite{nikravesh2007evolution}. Moreover, common sense provides a way to get solutions to problems with incomplete or imprecise information whereas classical logic is better to resolve problems well defined. Furthermore, if we always used classical logic, we would take only a few decisions throughout the day. 

While classical logic deals with binary sets of values, 0 (false) and 1 (true), fuzzy logic deals with a range of values that represents different degrees of truth on a scale between completely false and completely true \cite{Cueva-Fernandez2015fuzzy}. A range of values allows having more states and thus, making decisions knowing more information. For instance, fuzzy logic is able to optimise the moment when collected data should be pushed to a server \cite{Cueva-Fernandez2015fuzzy}. Applying classical logic, we would have only two states, push or wait, whereas if we applied fuzzy logic, we could have many states which would indicate the best moment to push data to the server and what data should be pushed.

Our daily life could be benefited with the combination of the IoT and fuzzy logic. For example, local businesses could enhance the management of temperature systems that involve the use of heating systems and air-conditioning systems in order to adapt the indoor conditions taking into consideration the outdoor conditions. Following this path, we would be able to achieve a really Smart City where the indoor locations will be capable of adapting themselves to the outdoor conditions. Using fuzzy logic will allow us to decide the best moment to heat up or to cool down the environment considering the outdoor conditions.

Moreover, fuzzy logic could help to reduce the energy consumption of cities' buildings in order to achieve the zero energy building (ZEBs). The ZEBs are a part of the way towards achieving the Smart Cities \cite{Kylili2015}.

In addition to use fuzzy logic in the IoT, there are also several platforms which allow not only publishing data gathered from our sensors but also consuming data gathered from third-party sensors that could help us to achieve our proposal.

The aim of this research is the development of a prototype that will combine fuzzy logic applied to the IoT and the use of IoT platforms in which the prototype where it will able to consume third-party data. The prototype will be able to automate a system that will control the environmental conditions of a specific place. Moreover, the prototype will address different issues in the world of the IoT like the energy savings and some points of the livability according to \cite{U.S.DepartmentofTransportation2015}. Our goals are the following:
\begin{itemize} 
\item Consuming data from online open IoT platforms.
\item Consuming data from sensors located in microcontrollers like Arduino or microcomputers like Raspberry Pi and working with actuators located in these devices.
\item Setting fuzzy rules to achieve a better performance in the suggested solution.
\item Keeping an optimum temperature inside a room for a long time.
\item Saving energy. Moreover, this can help to achieve economic savings.
\item Achieving some points of the livability.
\end{itemize}

The remainder of the paper is organised as follows. In Section~\ref{section:art}, we will introduce involved topics in this research like fuzzy logic, the IoT, IoT platforms, and the works related with this research. Section~\ref{section:case} will show the case of study, how it works, and how its architecture is. Section~\ref{section:sw-and-hw} will show the software and hardware used in this research. In Section~\ref{section:eval} we will cover the evaluation and discussion of the data obtained from comparing the behaviour of our approach with the behaviour of the traditional approach. Section~\ref{section:conclusions} will contain the conclusions of this paper and finally, in Section~\ref{section:future} we will describe the possible future work that can be done from here.

\section{State of the Art}
\label{section:art}
The principal aim of this research is to use fuzzy logic in the IoT context. However, before talking about the case study, it is necessary to review the involved concepts. Moreover, inside the IoT context we want to review a special concept, the IoT platforms. These platforms were created to bring the IoT to more people, industry, and researches and thus, to increase the possibilities of the IoT. With these platforms we can intercommunicate objects which are separated by thousands of kilometres or located in places that we cannot imagine. Maybe behind the enemy lines in a battlefield, inside a nuclear plant, in a sealed room full of chemicals or radioactive products, or even in places where they cannot be reached after being set \cite{bulusu2001scalable, Akyildiz2002}.

The fuzzy logic is also a technology presents in many researches. The uses of this technology are multiples and it could be useful in a wide range of fields. For example, the fuzzy logic can be used in the health field achieving the simulation of an artificial pancreas that is able to regulate blood glucose levels in an automatically way \cite{grant2007new} or sending health data in the best moment without draining the batteries of the involved devices \cite{bhunia2014ihealth}.

In the context of industry, the combination of the IoT and fuzzy logic can help in many situations like controlling the products life cycle \cite{chen2013intelligent}, monitoring and detecting fires \cite{maksimovic2014developing}, and making decisions about when doing certain procedures in industries \cite{zinonos2014wireless}.

Thus, in this section we will talk about the concepts which are involved in this research: \nameref{subsection:fuzzy}, \nameref{subsection:iot}, and \nameref{subsection:platforms}.

\subsection{Fuzzy Logic}
\label{subsection:fuzzy}
The term fuzzy logic was introduced on 1965 by Lotfi Asker Zadeh as a way to deal with common sense problems. He introduced fuzzy sets firstly in \cite{zadeh1965fuzzy} and after, fuzzy logic was defined and presented in \cite{zadeh1975fuzzy}. Since then, many researches have addressed topics related with fuzzy logic. A frequent issue where fuzzy logic is usually applied is the battery saving. For instance, Larios et al. \cite{Larios2012} succeeded in locating a device avoiding many location errors and therefore, reaching a better accuracy. Besides, their approach allows decreasing the energy consumption of different elements like the GPS. Chamodrakas and Martakos \cite{Chamodrakas2012} proposed to use the fuzzy set representation method to select a network in order to be connected in an efficient way, with a low energy consumption, with a high Quality of Service (QoS), and with good performance. Bagchi \cite{Bagchi2011} used fuzzy logic to keep the quality of playback of multimedia streaming and to achieve improvements in the energy consumption. Cueva-Fernandez et al. \cite{Cueva-Fernandez2015fuzzy, Cueva-Fernandez2015voice} made two proposals about how to use fuzzy logic in vehicles. In \cite{Cueva-Fernandez2015fuzzy}, they proposed the improvement of information exchange between vehicles sensors and servers in order to save energy whereas in \cite{Cueva-Fernandez2015voice}, they proposed a system to create applications through the voice using fuzzy logic. All of these researches have in common the use of fuzzy logic to make difficult decisions without having clear options to consider.

Fuzzy logic emerged to resolve problems that classical logic cannot address. Classical logic can only deal with binary sets of values (0 or 1), but there are many contexts in which dealing with more values is required. Having more than two values allows having more states and thus, making decisions with more information. These states are usually called linguistic variables and they can represent characteristics like `size' whose values could be `small', `very small', `big', `very big', and so on. This approach tries to resolve questions like how big a building is. The answer to this question depends on individual cognition because not everyone would respond the same. For example, for a person who lives in the countryside, a building composed by 6 floors could be big, but for a person who lives in New York, the same building could be small in comparison with the buildings placed in New York.

There are many examples that use fuzzy logic because they need more than two values to represent their states. Grant \cite{grant2007new} took the advantages of fuzzy logic to propose a new approach to the diabetic control. 

Fuzzy logic allows dealing with vague and ambiguous data in order to make decisions like a person would do it. This logic uses controllers called `adaptive controllers' \cite{grant2007new} or `expert systems' \cite{russell1995artificial} based in a rules (e.g., if X and Y then Z) to mimic the human fuzzy thinking. These rules represent the knowledge that drives the expert systems to decide optimal decisions. The rules definition is complex and it requires the use of the linguistic variables since the human representation of the knowledge is fuzzy. The process of understanding what a linguistic variable like `small' means is named fuzzification \cite{portmann2010prometheus}. 

\subsection{The Internet of Things}
\label{subsection:iot}
In the last years, the interconnection between heterogeneous and ubiquitous objects has been one of the most important issues of researches and business. This interconnection is better known as the Internet of Things \cite{Atzori2010, LuTan2010}.

Moreover, the IoT is also one of the most important technologies for the future according to the ONU in 2005 \cite{Atzori2010} where the United Nations predicted a new era with more data traffic generated by objects than generated by people. Furthermore, in 2008, the United States National Intelligence Council reported that they considered the IoT as one of the six technologies in which the United States would be more interested in from 2008 to 2025 \cite{nic2008disruptive}.

However, there is not a standard or unique way to work with the IoT. There are many different platforms to interconnect objects through Internet \cite{GonzalezGarcia2014, GonzalezGarcia2014Congress} and a lot of standards to communicate the objects like Near Field Communications (NFC), Radio Frequency Identification (RFID), Bluetooth, or Message Queue Telemetry Transport (MQTT). 

The IoT is a set of technologies that can be used for many purposes. The use of sensors in combination with Smart Objects, also called Intelligent Products, and Not-Smart Objects \cite{GonzalezGarcia2017} like actuators or other sensors, allows us to read data from the environment and develop a wide range of solutions for many topics. In \cite{GonzalezGarcia2017}, authors do a review about these Smart and Not-Smart Objects, and introduce some applications that show the wide range of possibilities that the IoT, in combination with Smart and Not-Smart Objects, can offer. Examples of applications of the IoT could be the detection of drowsiness at the wheel through smart devices with many sensors like Smartwatches in order to prevent road accidents \cite{Rios-Aguilar2015}.

\subsection{IoT platforms}
\label{subsection:platforms}
Due to the popularity and the growth of the IoT, there have been developed several platforms to interconnect heterogeneous objects through the Internet. 

The heterogeneity is a reason for the development of IoT platforms. As it is said in \cite{IglesiasFeijoo2015}, the development of IoT solutions for different Smart Objects from several manufacturers is difficult because of the lack of standards like communication and interconnection protocols. Therefore, they propose a new development model based on handling the user interfaces on the Cloud. By this way, their proposal achieves the development of cross-platform and ubiquitous applications. However, this approach does not take into account the interconnection between objects as other IoT platforms do. There are many platforms and many types and their aims can be very different.

Several platforms were developed for business community like Xively \cite{xively} and they allow businesses to be benefited from the IoT advantages in order to connect their products with their users in a safe way, manage their information, improve customer satisfaction, and make new revenue sources.

Many others like ThingSpeak \cite{thingspeak} or Paraimpu \cite{paraimpu}, allow registered users to consume open data in many formats like JSON, CSV, or XML and allow users to share their data from different sources. Both are free but the registrations are controlled by invitations or they require be approved.

There is also an IoT platform called Midgar \cite{GonzalezGarcia2014, GonzalezGarcia2014Congress} which generates applications which connect heterogeneous and ubiquitous objects through the Internet in order to facilitate the use of the IoT to any people which do not have computer science knowledge.

In our proposal, we opted for using ThingSpeak and Midgar in order to obtain data from third-party solutions.

\subsubsection{ThingSpeak}
ThinkSpeak \cite{thingspeak} is an open IoT platform for interconnecting objects using web standards. Some of its services are data registration, processing, and distribution; location based services, many plugins, and others useful services. The first thing to do in order to create new objects or services is to choose the device type (Arduino, Netduino, ioBridge, or others) and give the access data (IP, port, subnet mask, and others). This step allows creating a new channel to be used to push the connected devices data, show it, consume it through HTTP requests, or download it in XML, JSON, or CSV format. Moreover, this platform also offers a data accessibility control.

\subsubsection{Midgar}
Midgar \cite{GonzalezGarcia2014, GonzalezGarcia2014Congress} uses a Domain-Specific Language (DSL) called Midgar Object Interconnection Specific Language (MOISL) to generate applications that interconnect heterogeneous and ubiquitous objects through the Internet. Midgar users can interconnect different objects using a graphic DSL in an easy and fast way. Through this DSL, users can define the application flow in order to connect many objects, read any data from their services, and invoke certain actions in the same objects or in other connected objects. These actions can be different among different objects and one application can interconnect many different objects and invoke different actions. Finally, Midgar processes the model that was defined by the users with MOISL, and creates the application which interconnect the different selected objects and send the different instructions in each circunstance, according to the users definition.

\subsection{Related work}
There are several commercial solutions that allow handling the temperature of specific places. Two of this solutions are Loxone \cite{loxone} and tado\degree \ \cite{tado}. Moreover, there are also some researches that combine the IoT and fuzzy logic like Vitruvius \cite{Cueva-Fernandez2015fuzzy, Cueva-Fernandez2015voice}

\subsubsection{Loxone} 
Loxone \cite{loxone} is a home automation system that can control blinds, lights, music systems, or climatisation systems. If we are focused in the temperature, this system has several advantages and disadvantages in comparison with our approach. Some advantages are that it takes into consideration the user geolocation and it registers the temperature to calculate trends and to improve the future work. However, it also has disadvantages like the high buying price and the installation complexity. Moreover, Loxone does not use third-party data whereas our purpose can gather data from online open IoT platforms.

\subsubsection{tado\degree}
tado\degree \ \cite{tado} is a smart thermostat that can be controlled through a web interface or with a smartphone. It can use the user geolocation like Loxone although its installation is easier than Loxone. tado\degree \ also allows us to save money with a good temperature control. Moreover, it generates detailed reports and it takes into consideration the weather forecast to keep the ambient temperature. The last feature is comparable with the gathering of data from online open IoT platforms so this is a point in common between tado\degree \ and our proposal. However, there are also disadvantages like the high buying price and it does not automate anything so it cannot be considered smart.

\subsubsection{Vitruvius}
Vitruvius \cite{Cueva-Fernandez2015fuzzy, Cueva-Fernandez2015voice} is a platform which allows users to generate applications in real time that use sensors installed in vehicles. For that purpose, Vitruvius has a web application composed by an application editor that allows designing the applications. Some of available sensors are speed sensors, temperature sensors, and many similar others. Vitruvius combines the IoT with fuzzy logic in order to reduce the push data frequency. Its first versions did not use fuzzy logic, therefore, the data were being pushed to server every second \cite{Cueva-Fernandez2014}. Due to fuzzy logic, the amount of data pushed were reduced and the data quality and the data analysis performance were improved. However, the use of fuzzy logic makes it harder to detect errors. Whereas Vitruvius used fuzzy logic to automate the pushing of data to server in the best moment to do it, our proposal use the fuzzy logic to automate systems that control the temperature with the intention of adjusting the temperature to obtain a good thermal sensation and thus, save money through the smart approach driven by fuzzy logic.

\section{Case study}
\label{section:case}
\begin{figure}
\centering
  \includegraphics[width=0.7\textwidth]{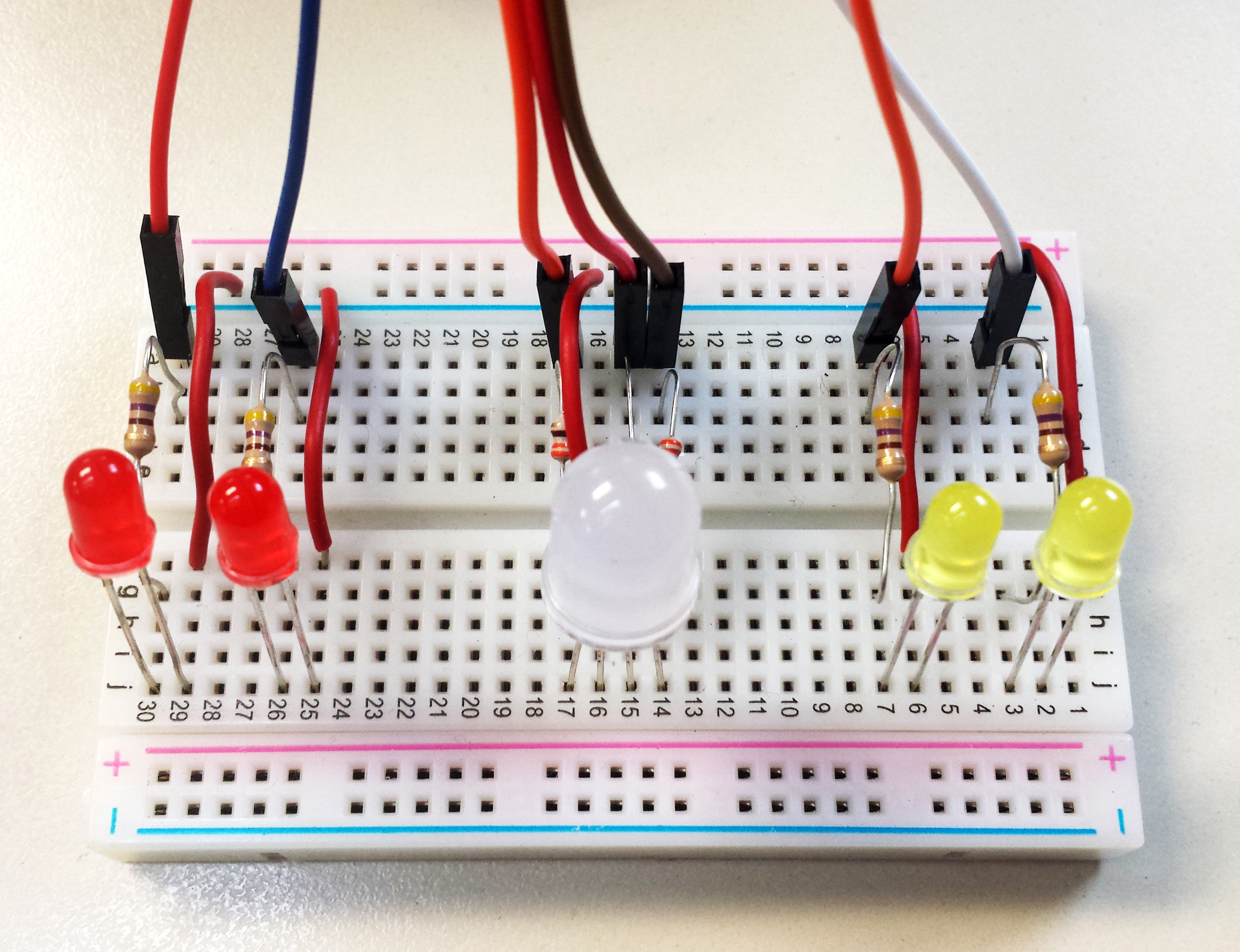}
  \caption{The five LEDs that simulate the temperature control.}
  \label{fig:leds}	
\end{figure}

In this paper we propose a system that controls the temperature of a specific place in an automated way using fuzzy logic. For this purpose, we developed a prototype inspired in the IoT context. The prototype consumes data from sensors and IoT platforms. The sensors were installed in microcontrollers and microcomputers. However, the microcomputers were used not only for consuming data but also for handling the actuators that simulate the temperature control. For this paper, we simulated the temperature control with five LEDs (Figure~\ref{fig:leds}) that represent the next five states in a closed room:
\begin{itemize}
\item The temperature would be extremely high and the air-conditioning systems would be on at maximum power. This state would be represented by two yellow LEDs.
\item The temperature would be high and the air-conditioning systems would be on at normal power. This state would be represented by one yellow led.
\item The temperature would be suitable and the heating systems and the air-conditioning systems would be off in order to keep the suitable temperature. This state would be represented by the blue led located in the middle.
\item The temperature would be low and the heating systems would be on at maximum power. This state would be represented by one red led.
\item The temperature would be extremely low and the heating systems would be on at maximum power. This state would be represented by two red LEDs.
\end{itemize}

\subsection{Temperature control}
\label{section:temp-control}
The developed prototype uses three input values from different IoT sources to make the decisions that control the temperature. 
\begin{itemize}
\item Outdoor humidity.
\item Outdoor temperature.
\item Indoor temperature.
\end{itemize}

\begin{figure}
\centering
  \includegraphics[width=\textwidth]{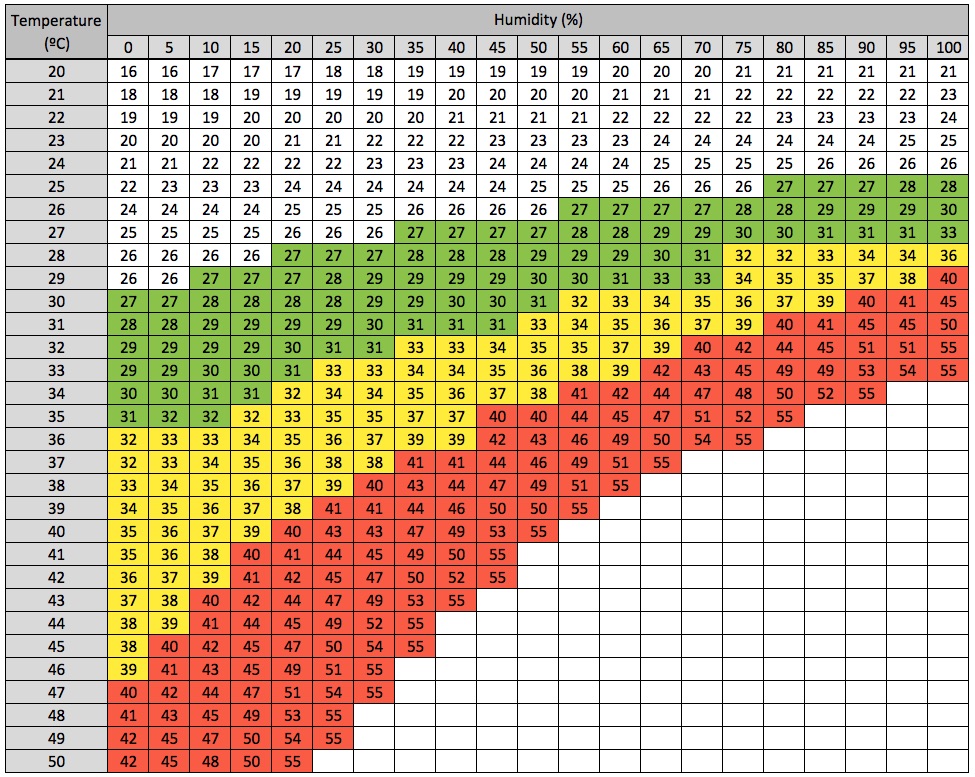}
  \caption{Apparent temperature according to the temperature and humidity.}
  \label{fig:temp-table}
\end{figure}

The prototype estimates the outdoor apparent temperature using the outdoor humidity, the outdoor temperature, and fuzzy rules. These fuzzy rules were based in the values of the table shown in Figure~\ref{fig:temp-table} whose input values are the temperature and the humidity. The output after applying the fuzzy rules are a set of values that indicate the degree of truth of seven linguistic variables that were used in order to represent the apparent temperature. The linguistic variables used were the next fuzzy sets:
\begin{itemize}
\item Extremely low: Between -15\degree C and -5\degree C.
\item Very low: Between -7\degree C and 3\degree C.
\item Low: Between 0\degree C and 18\degree C.
\item Normal: Between 14\degree C and 24\degree C.
\item High: Between 22\degree C and 30\degree C.
\item Very high: Between 28\degree C and 38\degree C.
\item Extremely high: Between 35\degree C and 50\degree C.
\end{itemize}

The humidity and the outdoor temperature were also defined using linguistic variables in order to apply fuzzy logic. The outdoor temperature linguistic variables that we used were the same as we used to represent the apparent temperature. Nevertheless, the humidity linguistic variables were the next fuzzy sets:
\begin{itemize}
\item Very low: Less than 40\%.
\item Low: Between 30\% and 60\%.
\item Normal: Between 50\% and 70\%.
\item High: Between 60\% and 80\%.
\item Very high: More than 75\%.
\end{itemize}

\begin{figure}
\centering
  \includegraphics[width=0.9\textwidth]{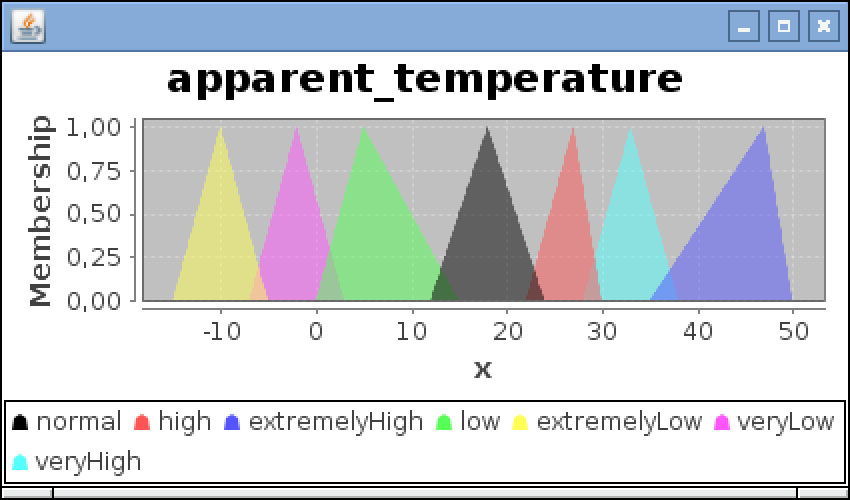}
  \caption{Distribution of linguistic variables for outdoor apparent temperature.}
  \label{fig:apparent-temp}
\end{figure}

\begin{figure}
\centering
  \includegraphics[width=0.9\textwidth]{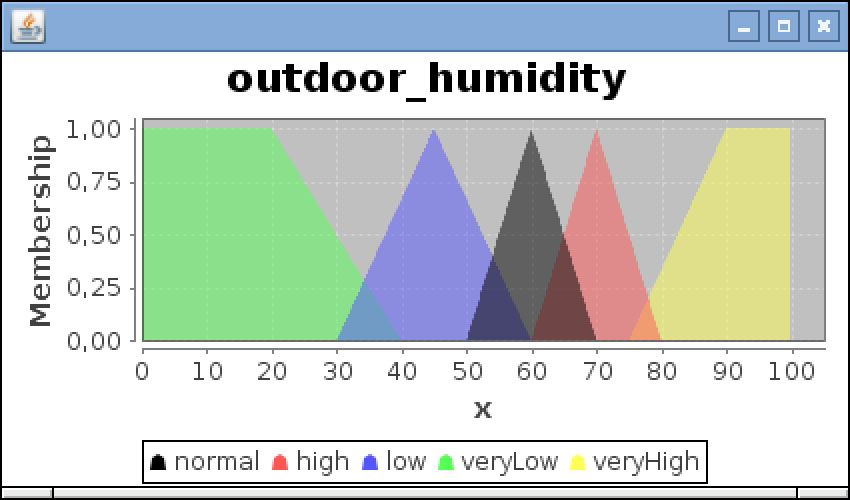}
  \caption{Distribution of linguistic variables for outdoor humidity.}
  \label{fig:humidity}
\end{figure}

From the degree of truth of each linguistic variable we select the centre of gravity in order to take the inputs of the next step. We can see the distribution of the linguistic variables for the outdoor apparent temperature in Figure~\ref{fig:apparent-temp} and for the outdoor humidity in Figure~\ref{fig:humidity}.

The outdoor humidity was obtained from an IoT platform that we talked about before, ThingSpeak, and the chosen format was JSON. The outdoor temperature was obtained from a Arduino Uno microcontroller with a temperature sensor connected to the Midgar platform, which was publishing the temperature data in JSON format. The indoor temperature was obtained from a temperature sensor located in a Raspberry Pi 2, where the prototype was running. The indoor temperature linguistic variables were the same as the ones used with the previous temperatures.

After estimating the outdoor apparent temperature, the prototype executes other fuzzy rules whose input is the indoor temperature centre of gravity and the outdoor apparent temperature centre of gravity in order to determine how the actuators should work. The output of these rules are the next linguistic variables and fuzzy sets and it defined the five states that we already mentioned before:
\begin{itemize}
\item Heating system at maximum power: Less than 15.
\item Heating system at normal power: Between 10 and 20.
\item No system running: Between 18 and 22.
\item Air condition system at normal power: Between 20 and 30.
\item Air condition system at maximum power: More than 25.
\end{itemize}

After calculating the centre of gravity, we obtained the action to do. All values shown in this section were influenced due to the experience obtained during the research and they should be adjusted to the conditions of each place.

\subsection{Architecture}
\begin{figure}
\centering
  \includegraphics[width=0.9\textwidth]{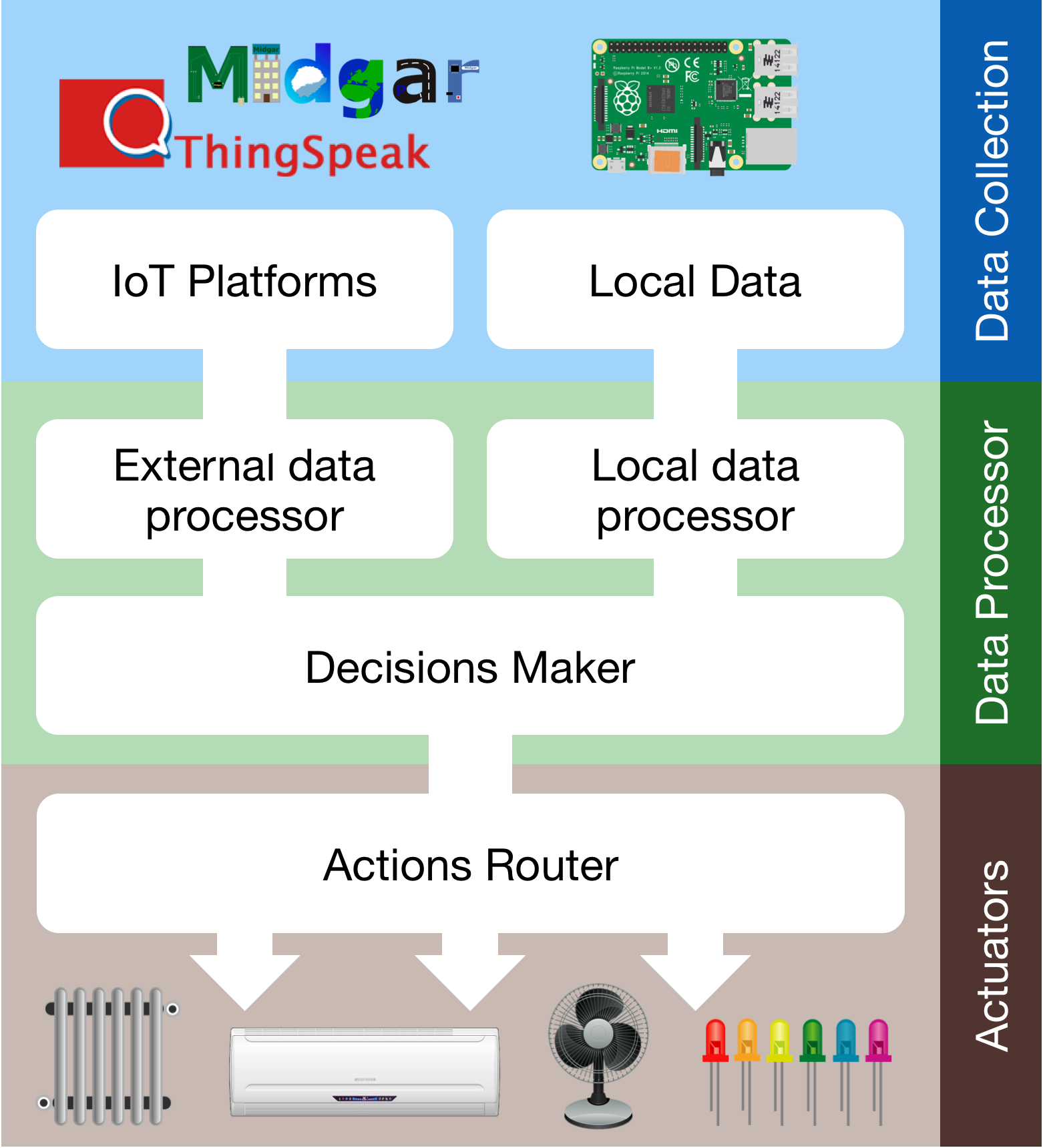}
  \caption{System architecture.}
  \label{fig:architecture}
\end{figure}

The system’s architecture can be divided in the next three layers: \textbf{Data collection}, \textbf{Data processor}, and \textbf{Actuators}. The Figure~\ref{fig:architecture} shows the three layers and their modules. Each layer depends on the others. The first one, \textbf{Data collections}, contains the modules which gather data to be processed from external and local sources. The layer \textbf{Data processor} takes the collected data and processes them in order to make decisions. The last layer, \textbf{Actuators}, takes the decisions made in the previous layer and routes them to the target actuator in order to do the appropriate action.

\subsubsection{Data collection}
Our approach requires data from different IoT platforms in order to handle variety data from external sources. The system is prepared to use any platform with minor changes. In our tests we focus in the use of two IoT platforms, one for temperature, Midgar, and another for humidity, ThingSpeak.

The IoT platform used to collect the temperature data was Midgar. Midgar is useful to interconnect Smart Objects in an easy way. In fact, knowledge about languages programming is not necessary \cite{GonzalezGarcia2014, GonzalezGarcia2014Congress}. We used Midgar to connect an Arduino Uno with our system through HTTP requests using a REST web service. This Arduino had a temperature sensor that we used to obtain the outdoor temperature.

In order to collect the humidity data, we used another IoT platform, ThingSpeak. This platform allows us to gather the humidity data through HTTP, that other users of the platform had published before.

Our system collects data not only from external sources but also from local sources. We used a Raspberry Pi 2 with a temperature sensor to obtain the temperature of the room where we want to handle the heating and air condition systems.

The data obtained are preprocessed in order to transform it in a common format that the Data processor can handle.

\subsubsection{Data processor}
\label{subsubsection:data_processor}
The output of the previous layer is a set of properly formatted values that represents the outdoor temperature, the outdoor humidity, and the indoor temperature. These data have to be processed in order to decide what actions to do. For this purpose, we implemented two modules that process the external data and the local data. The external data sources are IoT platforms whereas local data sources are devices located in the room where we want handle the temperature.

In this layer, we convert the concrete values of temperatures and humidity in fuzzy values using the fuzzy sets and linguistic variables that it saw in Section~\ref{section:temp-control}. 

The external processor has fuzzy rules defined to be applied over the fuzzy values previously obtained, which represent the outdoor temperature and humidity. These fuzzy rules are shown in Table~\ref{table:apparent-temperature-rules}. With this fuzzy rules we obtain the degree of truth of each possible linguistic variable for the apparent temperature.

\begin{table}[]
\centering
\resizebox{\textwidth}{!}{
\begin{tabular}{ccccccccc}
\toprule
 &  & \multicolumn{7}{c}{\textbf{Outdoor temperature}} \\ \cline{3-9} 
 & \multicolumn{1}{c|}{\textbf{AND}} & \multicolumn{1}{c|}{\textbf{\begin{tabular}[c]{@{}c@{}}Extremely\\ Low\end{tabular}}} & \multicolumn{1}{c|}{\textbf{Very Low}} & \multicolumn{1}{c|}{\textbf{Low}} & \multicolumn{1}{c|}{\textbf{Normal}} & \multicolumn{1}{c|}{\textbf{High}} & \multicolumn{1}{c|}{\textbf{Very High}} & \textbf{\begin{tabular}[c]{@{}c@{}}Extremely\\ High\end{tabular}} \\ \cline{2-9} 
\multicolumn{1}{c|}{} & \multicolumn{1}{c|}{\textbf{Very Low}} & \multicolumn{1}{c|}{\begin{tabular}[c]{@{}c@{}}Extremely\\ Low\end{tabular}} & \multicolumn{1}{c|}{Very Low} & \multicolumn{1}{c|}{Very Low} & \multicolumn{1}{c|}{Normal} & \multicolumn{1}{c|}{Normal} & \multicolumn{1}{c|}{High} & \begin{tabular}[c]{@{}c@{}}Very High\end{tabular} \\ \cline{2-9} 
\multicolumn{1}{c|}{\multirow{4}{*}{\rotatebox{90}{\textbf{Outdoor humidity}}}} & \multicolumn{1}{c|}{\textbf{Low}} & \multicolumn{1}{c|}{\begin{tabular}[c]{@{}c@{}}Extremely\\ Low\end{tabular}} & \multicolumn{1}{c|}{Very Low} & \multicolumn{1}{c|}{Low} & \multicolumn{1}{c|}{Normal} & \multicolumn{1}{c|}{High} & \multicolumn{1}{c|}{Very High} & \begin{tabular}[c]{@{}c@{}}Extremely\\ High\end{tabular} \\ \cline{2-9} 
\multicolumn{1}{c|}{} & \multicolumn{1}{c|}{\textbf{Normal}} & \multicolumn{1}{c|}{\begin{tabular}[c]{@{}c@{}}Extremely\\ Low\end{tabular}} & \multicolumn{1}{c|}{Very Low} & \multicolumn{1}{c|}{Low} & \multicolumn{1}{c|}{Normal} & \multicolumn{1}{c|}{Very High} & \multicolumn{1}{c|}{\begin{tabular}[c]{@{}c@{}}Extremely\\ High\end{tabular}} & \begin{tabular}[c]{@{}c@{}}Extremely\\ High\end{tabular} \\ \cline{2-9} 
\multicolumn{1}{c|}{} & \multicolumn{1}{c|}{\textbf{High}} & \multicolumn{1}{c|}{Very Low} & \multicolumn{1}{c|}{Low} & \multicolumn{1}{c|}{Low} & \multicolumn{1}{c|}{Normal} & \multicolumn{1}{c|}{Very High} & \multicolumn{1}{c|}{\begin{tabular}[c]{@{}c@{}}Extremely\\ High\end{tabular}} & \begin{tabular}[c]{@{}c@{}}Extremely\\ High\end{tabular} \\ \cline{2-9} 
\multicolumn{1}{c|}{} & \multicolumn{1}{c|}{\textbf{Very High}} & \multicolumn{1}{c|}{Low} & \multicolumn{1}{c|}{Low} & \multicolumn{1}{c|}{Normal} & \multicolumn{1}{c|}{Normal} & \multicolumn{1}{c|}{\begin{tabular}[c]{@{}c@{}}Extremely\\ High\end{tabular}} & \multicolumn{1}{c|}{\begin{tabular}[c]{@{}c@{}}Extremely\\ High\end{tabular}} & \begin{tabular}[c]{@{}c@{}}Extremely\\ High\end{tabular} \\ \bottomrule
\end{tabular}
}
\caption{Fuzzy rules used to calculate the degree of truth of the possible apparent temperature.}
\label{table:apparent-temperature-rules}
\end{table}

After applying the fuzzy rules and getting the centre of gravity we obtain a value for apparent temperature. With this value and the indoor temperature, the system is able to make decisions about what action to do in order to handle the room’s temperature. That is why, we apply another set of fuzzy rules, shown in Table~\ref{table:action-rules}, whose input are the apparent temperature and the indoor temperature and whose output is the degree of truth of the action to do.

\begin{table}[]
\centering
\resizebox{\textwidth}{!}{%
\begin{tabular}{ccccccccc}
\toprule
\textbf{} &  & \multicolumn{7}{c}{\textbf{Apparent temperature}} \\ \cline{3-9} 
 & \multicolumn{1}{c|}{\textbf{AND}} & \multicolumn{1}{c|}{\textbf{\begin{tabular}[c]{@{}c@{}}Extremely\\ Low\end{tabular}}} & \multicolumn{1}{c|}{\textbf{Very Low}} & \multicolumn{1}{c|}{\textbf{Low}} & \multicolumn{1}{c|}{\textbf{Normal}} & \multicolumn{1}{c|}{\textbf{High}} & \multicolumn{1}{c|}{\textbf{Very High}} & \textbf{\begin{tabular}[c]{@{}c@{}}Extremely\\ High\end{tabular}} \\ \cline{2-9} 
\multicolumn{1}{c|}{} & \multicolumn{1}{c|}{\textbf{\begin{tabular}[c]{@{}c@{}}Extremely\\ Low\end{tabular}}} & \multicolumn{1}{c|}{\begin{tabular}[c]{@{}c@{}}Heating sys. at\\ max. power\end{tabular}} & \multicolumn{1}{c|}{\begin{tabular}[c]{@{}c@{}}Heating sys. at\\ max. power\end{tabular}} & \multicolumn{1}{c|}{\begin{tabular}[c]{@{}c@{}}Heating sys. at\\ max. power\end{tabular}} & \multicolumn{1}{c|}{\begin{tabular}[c]{@{}c@{}}Heating sys. at\\ max. power\end{tabular}} & \multicolumn{1}{c|}{\begin{tabular}[c]{@{}c@{}}Heating sys. at\\ max. power\end{tabular}} & \multicolumn{1}{c|}{\begin{tabular}[c]{@{}c@{}}Heating sys. at \\ max. power\end{tabular}} & \begin{tabular}[c]{@{}c@{}}Heating sys. at\\ max. power\end{tabular} \\ \cline{2-9} 
\multicolumn{1}{c|}{\textbf{}} & \multicolumn{1}{c|}{\textbf{Very Low}} & \multicolumn{1}{c|}{\begin{tabular}[c]{@{}c@{}}Heating sys. at\\ max. power\end{tabular}} & \multicolumn{1}{c|}{\begin{tabular}[c]{@{}c@{}}Heating sys. at\\ max. power\end{tabular}} & \multicolumn{1}{c|}{\begin{tabular}[c]{@{}c@{}}Heating sys. at\\ max. power\end{tabular}} & \multicolumn{1}{c|}{\begin{tabular}[c]{@{}c@{}}Heating sys. at\\ max. power\end{tabular}} & \multicolumn{1}{c|}{\begin{tabular}[c]{@{}c@{}}Heating sys. at\\ max. power\end{tabular}} & \multicolumn{1}{c|}{\begin{tabular}[c]{@{}c@{}}Heating sys. at\\ max. power\end{tabular}} & \begin{tabular}[c]{@{}c@{}}Heating sys. at\\ max. power\end{tabular} \\ \cline{2-9} 
\multicolumn{1}{c|}{\multirow{5}{*}{\rotatebox{90}{\textbf{Indoor temperature}}}} & \multicolumn{1}{c|}{\textbf{Low}} & \multicolumn{1}{c|}{\begin{tabular}[c]{@{}c@{}}Heating sys. at\\ max. power\end{tabular}} & \multicolumn{1}{c|}{\begin{tabular}[c]{@{}c@{}}Heating sys. at\\ max power\end{tabular}} & \multicolumn{1}{c|}{\begin{tabular}[c]{@{}c@{}}Heating sys. at\\ max. power\end{tabular}} & \multicolumn{1}{c|}{\begin{tabular}[c]{@{}c@{}}Heating sys. at\\ normal power\end{tabular}} & \multicolumn{1}{c|}{\begin{tabular}[c]{@{}c@{}}Heating sys. at\\ normal power\end{tabular}} & \multicolumn{1}{c|}{\begin{tabular}[c]{@{}c@{}}Heating sys. at\\ normal power\end{tabular}} & \begin{tabular}[c]{@{}c@{}}Heating sys. at\\ normal power\end{tabular} \\ \cline{2-9} 
\multicolumn{1}{c|}{} & \multicolumn{1}{c|}{\textbf{Normal}} & \multicolumn{1}{c|}{\begin{tabular}[c]{@{}c@{}}Heating sys. at\\ normal power\end{tabular}} & \multicolumn{1}{c|}{\begin{tabular}[c]{@{}c@{}}Heating sys. at\\ normal power\end{tabular}} & \multicolumn{1}{c|}{\begin{tabular}[c]{@{}c@{}}Heating sys. at\\ normal power\end{tabular}} & \multicolumn{1}{c|}{\begin{tabular}[c]{@{}c@{}}No system\\ running\end{tabular}} & \multicolumn{1}{c|}{\begin{tabular}[c]{@{}c@{}}No system\\ running\end{tabular}} & \multicolumn{1}{c|}{\begin{tabular}[c]{@{}c@{}}No system\\ running\end{tabular}} & \begin{tabular}[c]{@{}c@{}}Air condition sys.\\ at normal power\end{tabular} \\ \cline{2-9} 
\multicolumn{1}{c|}{} & \multicolumn{1}{c|}{\textbf{High}} & \multicolumn{1}{c|}{\begin{tabular}[c]{@{}c@{}}No system\\ running\end{tabular}} & \multicolumn{1}{c|}{\begin{tabular}[c]{@{}c@{}}No system\\ running\end{tabular}} & \multicolumn{1}{c|}{\begin{tabular}[c]{@{}c@{}}No system \\ running\end{tabular}} & \multicolumn{1}{c|}{\begin{tabular}[c]{@{}c@{}}No system\\ running\end{tabular}} & \multicolumn{1}{c|}{\begin{tabular}[c]{@{}c@{}}Air condition sys.\\ at normal power\end{tabular}} & \multicolumn{1}{c|}{\begin{tabular}[c]{@{}c@{}}Air condition sys.\\ at normal power\end{tabular}} & \begin{tabular}[c]{@{}c@{}}Air condition sys.\\ at max. power\end{tabular} \\ \cline{2-9} 
\multicolumn{1}{c|}{} & \multicolumn{1}{c|}{\textbf{Very High}} & \multicolumn{1}{c|}{\begin{tabular}[c]{@{}c@{}}Air condition sys.\\ at normal power\end{tabular}} & \multicolumn{1}{c|}{\begin{tabular}[c]{@{}c@{}}Air condition sys.\\ at normal power\end{tabular}} & \multicolumn{1}{c|}{\begin{tabular}[c]{@{}c@{}}Air condition sys.\\ at normal power\end{tabular}} & \multicolumn{1}{c|}{\begin{tabular}[c]{@{}c@{}}Air condition sys.\\ at normal power\end{tabular}} & \multicolumn{1}{c|}{\begin{tabular}[c]{@{}c@{}}Air condition sys.\\ at max. power\end{tabular}} & \multicolumn{1}{c|}{\begin{tabular}[c]{@{}c@{}}Air condition sys.\\ at max. power\end{tabular}} & \begin{tabular}[c]{@{}c@{}}Air condition sys.\\ at max. power\end{tabular} \\ \cline{2-9} 
\multicolumn{1}{c|}{} & \multicolumn{1}{c|}{\textbf{\begin{tabular}[c]{@{}c@{}}Extremely\\ High\end{tabular}}} & \multicolumn{1}{c|}{\begin{tabular}[c]{@{}c@{}}Air condition sys.\\ at max. power\end{tabular}} & \multicolumn{1}{c|}{\begin{tabular}[c]{@{}c@{}}Air condition sys.\\ at max. power\end{tabular}} & \multicolumn{1}{c|}{\begin{tabular}[c]{@{}c@{}}Air condition sys.\\ at max. power\end{tabular}} & \multicolumn{1}{c|}{\begin{tabular}[c]{@{}c@{}}Air condition sys.\\ at max. power\end{tabular}} & \multicolumn{1}{c|}{\begin{tabular}[c]{@{}c@{}}Air condition sys.\\ at max. power\end{tabular}} & \multicolumn{1}{c|}{\begin{tabular}[c]{@{}c@{}}Air condition sys.\\ at max. power\end{tabular}} & \begin{tabular}[c]{@{}c@{}}Air condition sys.\\ at max. power\end{tabular} \\ \bottomrule
\end{tabular}
}
\caption{Fuzzy rules used to calculate the degree of truth of possible actions to do.}
\label{table:action-rules}
\end{table}

The made decisions are send to the next layer in order to do the corresponding action with the corresponding actuator.  

\subsubsection{Actuators}
The last layer is composed by the actuators that are connected like the heating systems and the air condition systems, although more actuators could be available like fans, LEDs, thermostats, or any others.

The aim of this layer is to route the actions to the corresponding actuators. For example, if the decision made in the \nameref{subsubsection:data_processor} layer was switching on the heating system, the Actuators layer should have switched on the heating system.

\section{Used software and hardware}
\label{section:sw-and-hw}
In order to develop the proposed solution by this paper, we needed the use of different types of software and hardware components.

The developed system’s language is Java 8 and it uses the libraries jFuzzyLogic\footnote{jFuzzyLogic: http://jfuzzylogic.sourceforge.net/html/index.html} \cite{cingolanijfuzzylogic, cingolani2012jfuzzylogic}, Pi4J 1.0\footnote{Pi4J: http://pi4j.com}, and Gson 2.3.1\footnote{Gson: https://github.com/google/gson}.

The used hardware was an Arduino UNO with a TMP36 temperature sensor and a Raspberry Pi 2 with several electronic components connected to its GPIOs:
\begin{itemize}
\item TMP36 temperature sensor.
\item MCP3008 analogic-to-digital converter.
\item 2 yellow LEDs, 2 red LEDs, and 1 RGB LED.
\item Several resistors.
\end{itemize}

\section{Evaluation and discussion}
\label{section:eval}
This paper proposes a system to control the temperature in an automated and efficient way. Traditional thermostats control the temperature switching on and switching off the heating system or the air condition system when the temperature is higher or lower than a specific value which is set manually. For instance, a thermostat could be set to switch on the heating system when the temperature is lower than 20\degree C. The problem is when the temperature is changing near 20\degree C (e.g., between 19\degree C and 21\degree C), because the system would switch on and switch off constantly. The use of fuzzy sets and linguistic variables allows us to determine ranges in which the temperature is suitable. Moreover, we are able to determine the best moment when heating systems and air condition systems have to be switched on and switched off taking the outdoor ambient conditions and using fuzzy logic.

In order to illustrate this, we made an evaluation of our approach where we compared the times that a heating system and an air condition system was switched on and switched off using an old thermostat with using our new fuzzy approach that also takes into account the outdoor conditions and the indoor temperature. In Table~\ref{tab:temps}, it is shown the outdoor temperature, humidity, and the indoor temperature that we used to do the simulation in this evaluation.

\begin{table}[]
\centering
\begin{tabular}{@{}cccc@{}}
\toprule
Time  & Humidity & Temp. Outdoor & T. Indoor \\ \midrule
00:00 & 66       & 15            & 18        \\
01:00 & 69       & 14            & 17        \\
02:00 & 73       & 12            & 16        \\
03:00 & 75       & 11            & 15        \\
04:00 & 76       & 10            & 15        \\
05:00 & 78       & 9             & 14        \\
06:00 & 76       & 9             & 14        \\
07:00 & 74       & 9             & 14        \\
08:00 & 72       & 10            & 14        \\
09:00 & 62       & 12            & 15        \\
10:00 & 53       & 15            & 15        \\
11:00 & 44       & 17            & 15        \\
12:00 & 38       & 19            & 18        \\
13:00 & 33       & 21            & 20        \\
14:00 & 28       & 23            & 23        \\
15:00 & 28       & 24            & 23        \\
16:00 & 29       & 25            & 24        \\
17:00 & 30       & 26            & 26        \\
18:00 & 36       & 27            & 26        \\
19:00 & 40       & 27            & 26        \\
20:00 & 42       & 28            & 26        \\
21:00 & 33       & 25            & 23        \\
22:00 & 30       & 23            & 22        \\
23:00 & 34       & 20            & 20        \\ \bottomrule
\end{tabular}
\caption{Ambient conditions of a full day.}
\label{tab:temps}
\end{table}

\begin{figure}
\centering
  \includegraphics[width=0.9\textwidth]{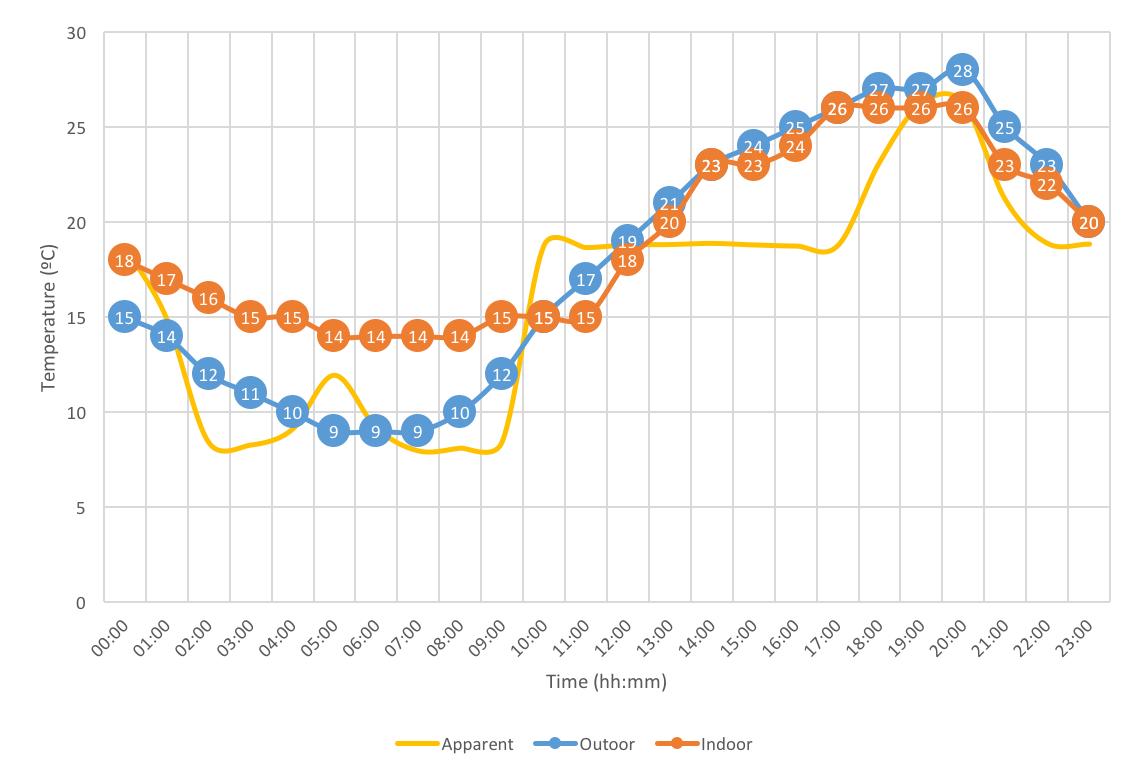}
  \caption{The obtained temperatures for the evaluation.}
  \label{fig:temperatures}
\end{figure}

We calculated the apparent temperature using the outdoor temperature and humidity, in order to use it with the indoor temperature to determine the best moment to switch on and to switch off the heating system and the air condition system. In Figure~\ref{fig:temperatures} we can see the three temperatures used in the evaluation. The blue curve is the outdoor temperature, the yellow curve is the apparent temperature calculated from outdoor temperature and humidity, and the orange curve is the temperature indoor during a full day without using any system to control the temperature. The apparent temperature and the outdoor temperature are different because the humidity was variable and the apparent temperature is calculated from outdoor temperature and humidity.

The old heating system that we simulate in this evaluation had two heating stages and the old air condition system had two airflow stages to control the temperature. The configuration used is the next:
\begin{itemize}
\item Heating system at maximum power: Below 15\degree C.
\item Heating system at normal power: Between 15\degree C and 17 \degree C.
\item Air condition system at maximum power: Above 25\degree C.
\item Air condition system at normal power: Between 23\degree C and 25\degree C.
\item Both systems switched off: Between 18\degree C and 22 \degree C.
\end{itemize}

The new fuzzy systems used another configuration based in ranges instead of single values. The used ranges are the same as the fuzzy sets shown in Section~\ref{section:temp-control}.

\begin{figure}
\centering
  \includegraphics[width=0.9\textwidth]{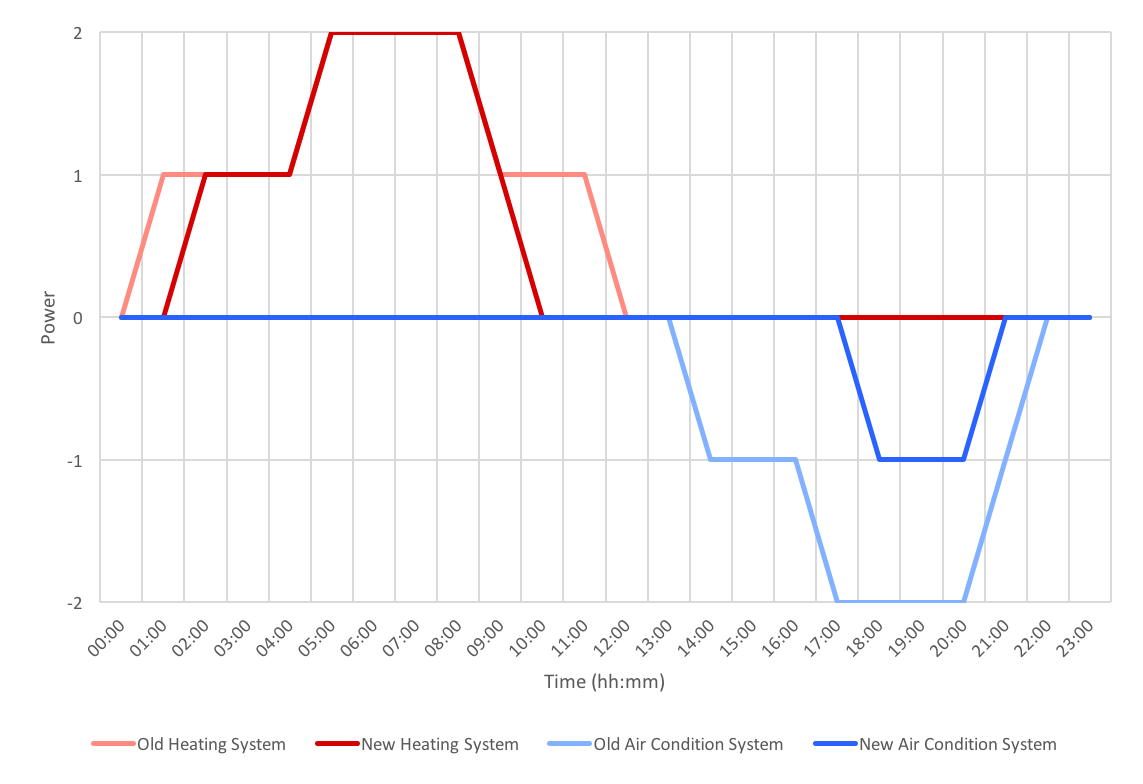}
  \caption{Stages of each system for one day.}
  \label{fig:power}
\end{figure}

After configuring the systems with the obtained and calculated values, we simulated what would have happened if these systems were used in the same conditions. In Figure~\ref{fig:power}, we can see when the systems switched on and when they switched off. The old heating system switched on one hour before our approach and it also switched off one hour later. Moreover, our approach for air condition system did not use the maximum power whereas the old system was at maximum power during 4 hours. Also, the new air condition system started working 4 hours later and stopped working 1 hour before.

\begin{figure}
\centering
  \includegraphics[width=0.9\textwidth]{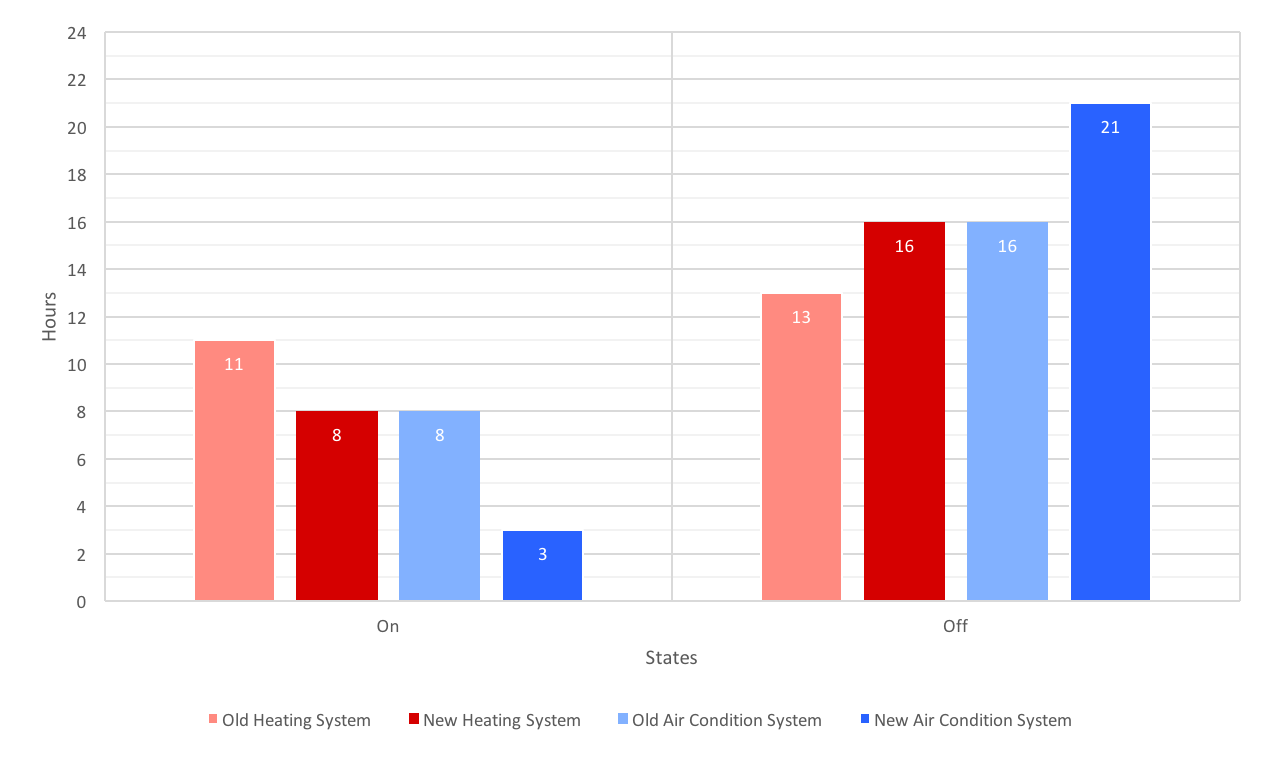}
  \caption{The number of hours per state.}
  \label{fig:times}
\end{figure}

Figure~\ref{fig:power} shows how the stages changed during a day and Figure~\ref{fig:times} shows how many hours each system was switched on and switched off. We can see that our approach was off for more hours. 

Firstly, the old heating system was on for 11 hours whereas the new heating system was on for 8 hours. There were 3 hours or approximately 27.27\% of energy saving. And secondly, the old air condition system was on for 8 hours whereas the new air condition system was off for 21 hours so there were 5 hours or approximately 62.5\% of energy saving. Therefore, we can assume that there was a certain saving energy. 

These results must be interpreted carefully. In spite of defining a temperature ranges, the system did not switch on whenever the temperature was out of range. Therefore, this approach may not be used in context of critical situations when the control of temperature cannot have margin of errors. The explanation of this behaviour could be the use of external conditions. Considering external conditions may be a good idea for the temperature control in rooms where there are people or objects that need specific environmental conditions. It could be a way of adapting the room for people who are entering and leaving the room. However, there are a lot of situations that this approach could be used.

\section{Conclusions}
\label{section:conclusions}
The future of computers is related with the IoT and each progress is very important for the development of technologies from the IoT context. In this paper we presented our approach about the integration of fuzzy logic into the IoT context.

We have proposed the use of fuzzy logic in order to control the temperature of a specific room taking into consideration the outdoor conditions, specifically, the outdoor temperature and humidity to obtain the outdoor apparent temperature.

In order to obtain the outdoor data, we required devices that analyse the environment. These devices can be part of a Smart City or Smart Town, or they can be part of a set of nearby Smart Cities and Smart Towns.

With this data, the developed prototype simulates the control of the heating system and the air condition system so that they could be switched on and switched off considering the combination of outdoor conditions and indoor temperature.

In the evaluation of our proposal, we realised that using fuzzy logic to help systems that control the temperature allows us to save energy and thus, save money. For instance, we achieve an energy saving of around 40\% if we understand that time saving implies energy saving. However, fuzzy logic causes that the range in which the systems are off was too broad. Therefore, this approach would not be valid in the context of critical situations where the temperature would be an accurate value due to the consideration of external conditions in addition to local conditions.

\section{Future work}
\label{section:future}
The aim of this research was to reach a first approach that combines IoT technologies and fuzzy logic in order to improve the temperature control in specific locations, and from here, there are much work to do in future researches.
\begin{itemize}
\item \textbf{Improving the fuzzy logic configuration}: The configuration used in this paper is based in our experience during the research progress. Improving this configuration would allow getting better results, saving more energy, and reaching a better temperature control.
\item \textbf{Saving old temperatures in order to know if the temperature is rising or it is decreasing}: The actual approach does not take into consideration indoor temperature history. Knowing the previous temperatures helps the system to make better decisions. The system behaviour may not be the same if the temperature is rising or if it is decreasing. Taking into consideration the temperature progress  could help avoid changing the state constantly.
\item \textbf{Publishing the data in IoT platforms}: In this research we obtained the outdoor apparent temperature combining the outdoor temperature and humidity. The apparent temperature is a valued data and publishing it could be a good contribution.
\item \textbf{Adding more sensors}: In this paper we used three values, the outdoor temperature, the outdoor humidity, and the indoor temperature. Taking more values would allow us to improve the calculation of the apparent temperature. Moreover, we also may calculate the apparent indoor temperature which depends not only on humidity and temperature but also on the quantity of people in the room and many other factors.
\item \textbf{Creating a Domain-Specific Language in order to personalise the rules and the fuzzy sets in an easy way}: The fuzzy logic configuration depends on the location and on the people who is going to use the system, thus, adding a way of personalising the configuration for people who does not have knowledge about programming is a great improvement of this research. Developing a Domain-Specific Language (DSL) is the best solution to reach this proposal. A DSL may allow any user with minimum knowledge about computers, to personalise the system configuration.
\end{itemize}

\section{Acknowledgements}
This work was performed by the `Ingenier\'ia Dirigida por Modelos MDE-RG' research group at the University of Oviedo under Contract No. FC-15-GRUPIN14-084 of the research project `Ingenier\'a Dirigida Por Modelos MDE-RG'. Project financed by PR Proyecto Plan Regional.

\section*{References}
\bibliography{bibliography}

\end{document}